\documentclass[10pt,twocolumn,letterpaper]{article}

\usepackage{cvpr}
\usepackage{times}
\usepackage{epsfig}
\usepackage{graphicx}
\usepackage{amsmath}
\usepackage{amssymb}
\usepackage{bbm}
\usepackage{multirow}
\usepackage{subcaption}
\usepackage{authblk}

\makeatletter
\renewcommand\AB@affilsepx{ \ \ \ \ \ \ \protect\Affilfont}
\makeatother

\usepackage[pagebackref=true,breaklinks=true,letterpaper=true,colorlinks,bookmarks=false]{hyperref}

\cvprfinalcopy 

\def\cvprPaperID{4526} 

\ifcvprfinal\fi
\begin{document}

\title{RSA: Randomized Simulation as Augmentation for Robust Human Action Recognition}

\author[1]{Yi Zhang\thanks{Indicates equal contributions.}}
\author[*2]{Xinyue Wei}
\author[1]{Weichao Qiu}
\author[1]{Zihao Xiao}
\author[1]{Gregory D. Hager}
\author[1]{Alan Yuille}
\affil[1]{Johns Hopkins University}
\affil[2]{Tongji University}
\affil[ ]{
\tt\small \{edwardz.amg,sarahwei0210,qiuwch,a904794043,alan.l.yuille\}@gmail.com, \quad{\tt\small hager@cs.jhu.edu}
}

\maketitle

\begin{abstract}

Despite the rapid growth in datasets for video activity, stable robust activity recognition with neural networks remains challenging. This is in large part due to the explosion of possible variation in video -- including lighting changes, object variation, movement variation, and changes in surrounding context. An alternative is to make use of simulation data, where all of these factors can be artificially controlled. 
In this paper, we propose the Randomized Simulation as Augmentation (RSA) framework which augments real-world training data with synthetic data to improve the robustness of action recognition networks. We generate large-scale synthetic datasets with randomized nuisance factors. We show that training with such extra data, when appropriately constrained, can significantly improve the performance of the state-of-the-art I3D networks or, conversely, reduce the number of labeled real videos needed to achieve good performance.  Experiments on two real-world datasets NTU RGB+D and VIRAT demonstrate the effectiveness of our method. 

\end{abstract}

\section{Introduction}

Recognizing human actions robustly from video is an important task in computer vision that has many real-world applications. The recent advances of action recognition are partially due to the creation of large-scale labeled video datasets~\cite{karpathy2014large, carreira2017quo, sigurdsson2016hollywood}. However, performance on action recognition benchmarks is still not as good as object recognition in static images. One reason for this inferior performance is the extremely high intra-class variation -- the same action looks very different when performed by different people in different environments while captured from different camera viewpoints~\cite{rahmani2014hopc, rahmani2017learning}.  Factors, such as lighting, appearance, and camera viewpoints, which should not affect recognizing an action are called \textit{nuisance factors}. When collecting real-world video data different nuisance factors are often entangled together, e.g. people often play soccer during the day on grass fields. Therefore, it is often difficult or impractical to collect and annotate a large real-world video dataset that sufficiently covers the space of all nuisance factors. Current action recognition benchmarks are far from capturing all the variations of a single action. Therefore, models trained on these datasets are prone to be biased towards intrinsic dataset-specific features.

\begin{figure}[t]
\begin{center}

\includegraphics[width=1.0\linewidth]{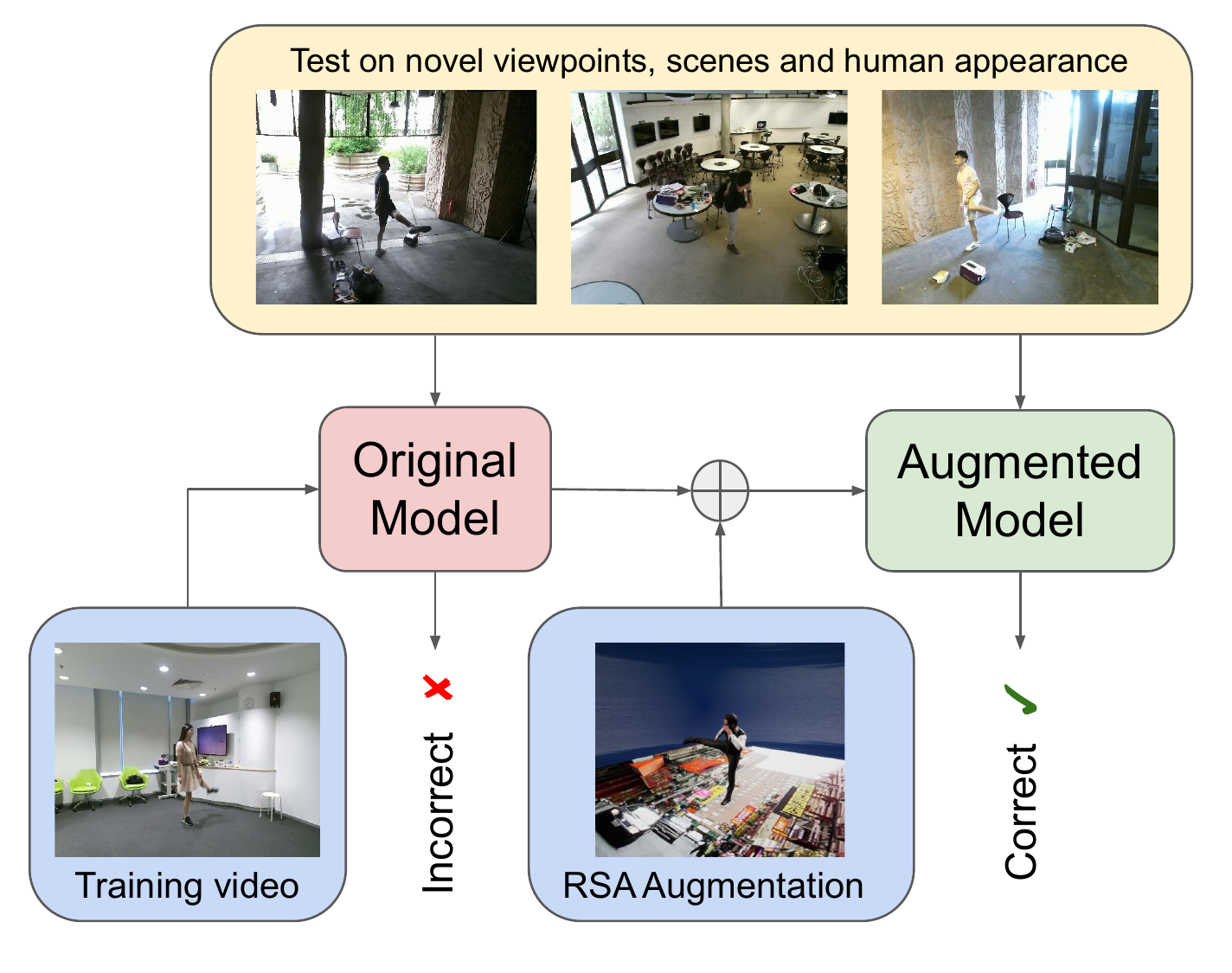}
\end{center}
   \caption{Illustration of our main idea. We use simulated videos to augment real training data to improve model robustness to nuisance factors, e.g. novel viewpoint, change of background and human appearance. }
\label{fig:teaser}
\end{figure}

In this paper we explore the possibility of augmenting real videos with simulated data to improve action recognition performance, particularly through robustness to nuisance factors, e.g. camera viewpoints, actor appearance and background. To this end, we propose the Randomized Simulation as Augmentation (RSA) framework in which we render actions in a virtual world where nuisance factors can be controlled. By varying the rendering parameters, we can generate a large number of action sequences that cover a large variation of nuisance factors. Adding these data to model training can reduce the possibility of overfitting and model bias. This is in the same spirit with domain randomization~\cite{sadeghi2016cad2rl,tremblay2018training,tobin2017domain,prakash2019structured}. 

Rendering synthetic action using computer graphics engines requires 3D human motion to drive 3D human models. For each action, we identify three possible ways to obtain the corresponding 3D human motion. First, we can search from existing large-scale motion capture libraries, i.e. CMU MoCap Library~\cite{cmumocap} and Human3.6M~\cite{h36m_pami}, which contain high-quality motions. A more scalable way is to obtain human motion from videos. Since 2D human pose estimation is relatively accurate, we can further obtain 3D motion from videos with depth information, i.e. RGB-D videos. Recently, monocular 3D human pose estimation becomes increasingly accurate. This supports extracting human motion directly from the RGB video. Furthermore, there is temporal consistency in videos which can be exploited to enhance pose estimation. 

Utilizing synthetic data requires dealing with the gap between synthetic and real domains, i.e. the reality gap. Since 1) renderers from computer graphics are not perfect and 2) it is often impractical to model every possible nuisance factors in the generation process, synthetic data will still have discrepancies from real data. This limits the potential for learning generalizable features from the synthetic domain. For deep models with large capacity, there is the additional complication that directly training with both real and simulated datasets jointly can result in learning models for the two domains separately~\cite{motiian2017unified}. We explore two possible solutions to this problem: 1) pretraining the network on synthetic data and then fine-tuning on real data; 2) performing domain adversarial training~\cite{ganin2016domain,tzeng2017adversarial} which minimizes the discrepancy of distributions of extracted features from two domains. We find that both methods outperform a baseline joint training method as well as performance from real data alone.

To support our methodology, we perform extensive experiments on two real-world datasets NTU RGB+D~\cite{Liu_2019_NTURGBD120} and VIRAT~\cite{oh2011large}. We show that augmentation with simulated data boosts the performance of the state-of-the-art action recognition model while improving the robustness to various nuisance factors. Meanwhile, we show that using synthetic data significantly reduce the number of labeled real videos needed to achieve good performance. We name our framework Randomized Simulation as Augmentation (RSA). 

Our contributions can be summarized as follows:
\begin{itemize}
    \item We propose an effective framework to augment real videos in with large-scale rendered synthetic videos to improve robustness to nuisance factors for human action recognition.
    \item We describe methods for effectively handling the domain gap between synthetic and real domains. 
    \item We demonstrate the effectiveness of our approach on two real-world datasets in terms of both improving performance of a state-of-the-art model and greatly reducing the need for labeled real data.
\end{itemize}

\begin{figure*}[t]
\begin{center}

  \includegraphics[width=\linewidth]{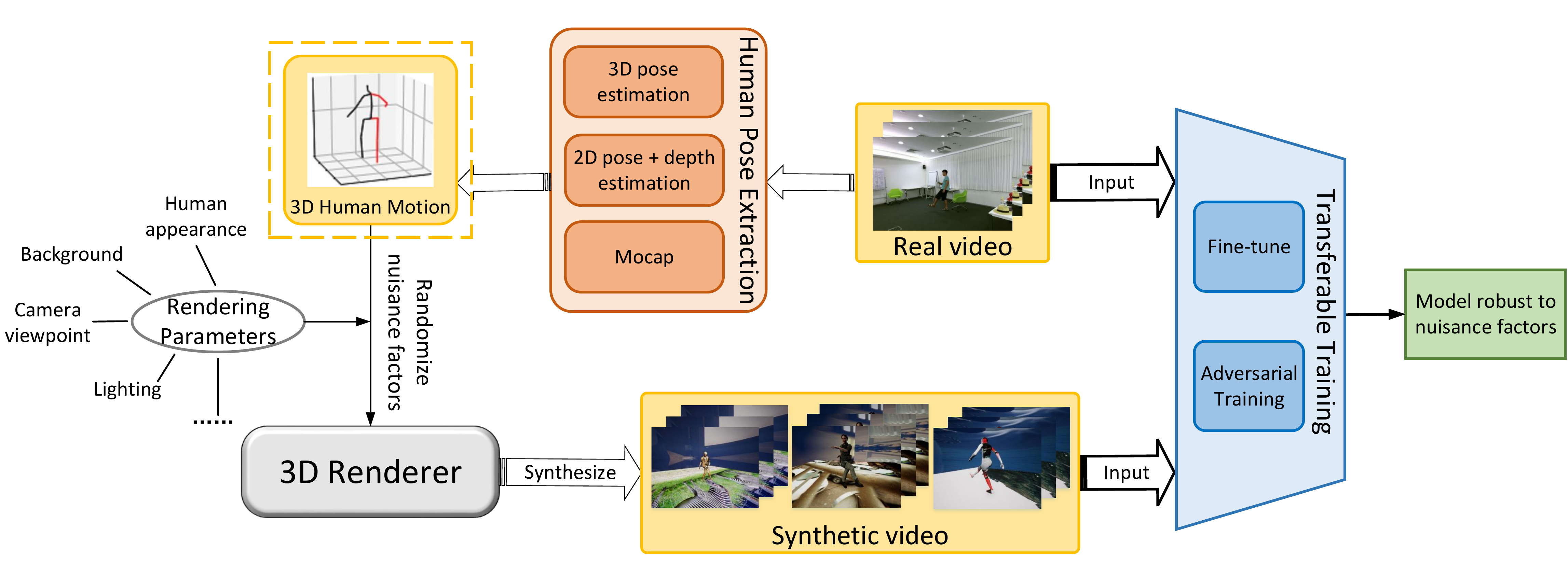}
\end{center}
   \caption{Overview of the RSA framework. Given real human action videos, we obtain corresponding 3D human motions for driving synthetic human models in a virtual environment. We render synthetic human action videos with randomized parameters that covers much more variations than the original dataset. Through transferable training with both real and synthetic data, the model achieves improved robustness to nuisance factors.}
\label{fig:main_idea}
\end{figure*}

\section{Related Work}

\textbf{Human Action Recognition.} Human action recognition has also benefited from the recent advances of deep convolutional neural networks (DCNN). Karpathy \textit{et al.}~\cite{karpathy2014large} demonstrated on a large dataset that DCNNs significantly improves action recognition accuracy over the feature-based baselines. Further improvements have been achieved through better temporal modeling~\cite{wang2016temporal, zhou2018temporal} and adding optical flow branch~\cite{simonyan2014two}. Another branch of methods for action recognition directly learn spatio-temporal features with 3D convolutions~\cite{tran2015learning, sun2015human}. Carreira and Zisserman~\cite{carreira2017quo} showed the effectiveness of inflating the filters of 2D CNNs pretrained on large-scale 2D image datasets as the initialization for 3D convolutional layers. To reduce the large computational cost of 3D convolutions, recent trends~\cite{qiu2017learning, tran2018closer, xie2018rethinking} have been focused on different ways to replace 3D convolutions with 2D convolutions.
Skeleton-based models~\cite{wang2016mining, soo2017interpretable, yan2018spatial} for human action recognition reason on dynamic human skeletons. Currently their performance is not on par with the RGB-based methods~\cite{yan2018spatial}. 

\textbf{Synthetic Data for Computer Vision.} Synthetic data has already been successfully applied in many computer vision tasks, such as stereo~\cite{mayer2016large}, optical flow~\cite{butler2012naturalistic, mayer2016large}, semantic segmentation~\cite{Richter_2016_ECCV, ros2016synthia}, human part segmentation~\cite{varol2017learning}, for which the groundtruth is either difficult or expensive to obtain in real world. 

Motivated by the fact that existing benchmarks lack the variety required to properly assess the performance of video analysis algorithms, Gaidon \textit{et al.}~\cite{gaidon2016virtual} propose to use synthetic data as a proxy to assess real-world performance on high-level computer vision tasks. The ability to control the nuisance factors in virtual world enables to study the robustness of vision algorithms~\cite{zhang2018unrealstereo}. Amlan \textit{et al.}~\cite{Kar_2019_ICCV} parameterizes synthetic dataset generator with a neural network and generates data to match content distributions for different tasks.
\cite{de2017procedural} is the first work to investigate using virtual worlds to generate synthetic videos for action recognition, where the author proposes to procedurally generate realistic action from a library of atomic motions.   Xavier\textit{et al.}~\cite{Puig_2018_CVPR} uses sequences of atomic actions and interactions to generate actions in a simulated household environment. Our work, however, goes beyond by randomizing the nuisance factors in synthetic data to improve model robustness and boost the performance in real domain.

\textbf{Domain Adaptation.} Computer vision models trained on a source domain often suffer from performance drop when tested on domains which have different marginal distributions. This phenomenon is known as domain shift which domain adaptation (DA) methods seek to alleviate. One type of DA methods directly minimizes a distance function, e.g. maximum mean discrepancy~\cite{long2015learning, long2017deep, ZellingerGLNS17} and CORAL function~\cite{sun2016deep}, for feature distributions of the source and target data. Domain adversarial training~\cite{ganin2016domain, tzeng2017adversarial} encourage to learn features which is indistinguishable between domains. Pixel-level DA methods~\cite{bousmalis2017unsupervised, hoffman2018cycada} align the distributions in the pixel space rather than the feature space.
For video domain adaptation, aligning the temporal dynamics between source and target domain is shown to be effective~\cite{sultani2014human}. For adaptation from synthetic to real domain, domain randomization~\cite{sadeghi2016cad2rl,tremblay2018training,tobin2017domain,prakash2019structured} is proposed which tries to randomize all irrelevant factors in virtual world forcing the model to focus on essential structure of the problem. 

\section{Method}



We present the Randomized Simulation as Augmentation (RSA) framework for robust action recognition in this section. We first provide the formulation of our method in a probabilistic perspective. We then introduce how we implement our framework in detail, followed by how we handle the domain gap between synthetic and real domain. 

\subsection{Formulation}

We use a simplified generative model, where an action is defined by human motion only, for generating human action videos while taking nuisance factors into consideration. Formally, we denote $x$ to be an RGB video. We formulate the generation of a human action video as a function $G(A, M, \{N_i\}_{i=1}^K)$, where random variable $A$ denotes the underlining action to generate, $M$ denotes human motion which contains a sequence of 3D human poses and $\{N_i\}_{i=1}^K$ are a set of pairwise independent random variables representing the nuisance factors, e.g. human appearance, surrounding environment and camera viewpoint, which is also independent of $A$ and $M$.  We obtain the following generative model for human action videos,
\begin{align}
\begin{split}
&P(x, A, M, \{N_i\}_{i=1}^K) = \\
& \mathbbm{1}\left[ x=G(A, M, \{N_i\}_{i=1}^K) \right] P(M|A)P(A) \prod_{i=1}^K P(N_i) 
\end{split}
\end{align}
where $\mathbbm{1}[\cdot]$ is indicator function. Therefore, the optimal discriminative recognizer is 
\begin{align}
\begin{split}
P(A|x) \propto \sum_{M} \sum_{N_1} \cdots \sum_{N_K} P(x, A, M, \{N_i\}_{i=1}^K)
\end{split}
\end{align}
which is our goal of approximation with a deep neural network and is not affected by nuisance factors. 

\subsection{Randomized Simulation as Augmentation for Robust Human Action Recognition}

It is costly and inefficient to sample enough data from the real $P(x, A, M, \{N_i\}_{i=1}^K)$, i.e. collect and annotate real videos, that sufficiently covers the variation of nuisance factors. We propose to use simulation in the 3D virtual world to augment more variation of nuisance factors. The whole framework is shown in Fig.~\ref{fig:main_idea}. The key idea is to use uniform distributions over all possible values for $\{N_i\}_{i=1}^K$. This is desirable because 1) the distributions of some nuisance factors should inherently be uniformly distributed, e.g. camera viewpoints; 2) even though other nuisance factors do have distributions that are not uniform in real world, e.g. the height of humans, such prior distribution should not be exploited too strongly by a robust recognition model. 

\begin{figure}[t]
\begin{center}
   \includegraphics[width=0.9\linewidth]{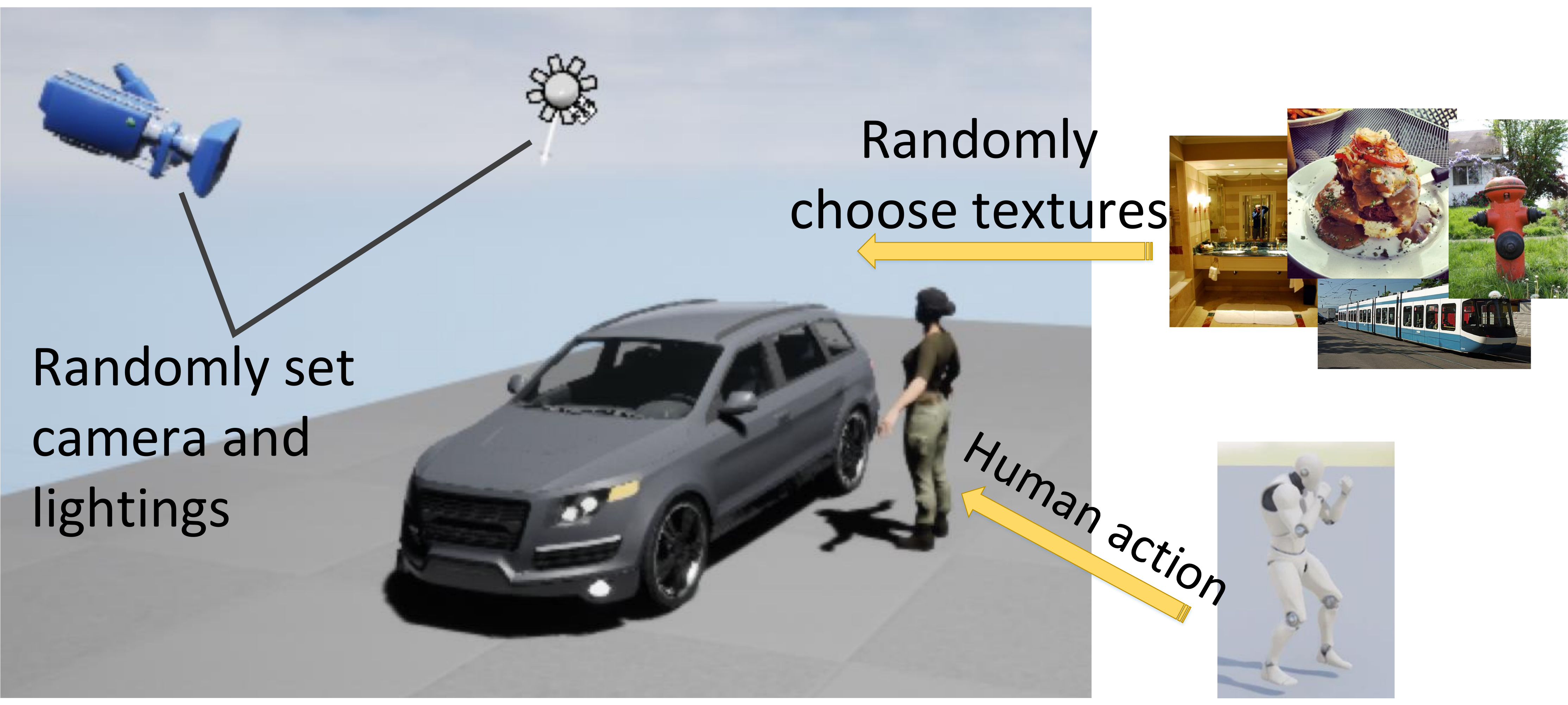}
\end{center}
   \caption{The virtual environment for generating synthetic data. A virtual camera can record the action from different viewpoint. The textures are given to sky, ground, vehicle and actor.}
\label{fig:camera}
\end{figure}

Specifically, we use a 3D renderer to approximate the real generator $G$. As it is shown in Fig.~\ref{fig:camera}, the renderer uses human motion sequences to drive 3D human skeletal meshes in a virtual environment. 

For the purposes of simulation, we need a motion model. 
We learn the model $P(M|A)$ from existing datasets. For action category $a$, let $\{(\hat{x}_i, a)\}_{i=1}^n$ denote the training examples from an existing dataset. The most direct way is to estimate 3D human motion $\hat{M}_i$ from $\hat{x}_i$, by which we obtain a motion set $\{(\hat{M}_i)\}_{i=1}^n$.  For datasets where RGB-D videos are provided, we can simply perform 2D human joint estimation which is relatively accurate and reconstruct the 3D pose with the help of depth information. We subsequently  assume $M$ takes value from $\{(\hat{M}_i)\}_{i=1}^n$ uniformly. Furthermore, we use action category name to query similar 3D human motion from existing large-scale motion capture libraries such as CMU MoCap Library~\cite{cmumocap}.

We consider three main nuisance factors -- human appearance, background environment and camera viewpoints -- that are controlled by rendering parameters to follow an approximately uniform distribution. We obtain a random human mesh from MakeHuman~\cite{makehuman} by random sampling from a set of parameters that define a humanoid. Unlike previous work~\cite{de2017procedural}, our method does not require realistic human texture. Instead, we use images randomly sampled from the MSCOCO dataset~\cite{lin2014microsoft} as textures. The human action is rendered in a simple virtual environment consisting of a floor and background sky sphere. Similarly, the textures of the floor and sky sphere are sampled from the same dataset. For the camera, we set the resolution to be $640\times480$ and an FOV of $90$ degrees. As shown in Fig.~\ref{fig:camera}, the camera always faces toward the human model and is controlled by three parameters, i.e. distance to the human model, azimuth and elevation, from which a random camera viewpoint is sampled for each video. 

\subsection{Transfer from Synthetic to Real}

Since the graphics renderer we utilized is only an approximation to the real world generator $G$, there are still discrepancies between real and synthetic action videos. Thus, a model trained with synthetic data will not generalize to real data. This problem is called the reality gap. Directly training networks on the mixture of synthetic and real data may not help the model to achieve better performance on the real domain. Because the huge capacity of deep networks enables learning to handle two domains separately as shown in Fig.~\ref{fig:feature_embed}. 

To overcome the reality gap, we investigate two strategies, i.e. finetuning and domain adversarial training~\cite{ganin2016domain, tzeng2017adversarial}. For finetuning, the action recognition network is pretrained on the synthetic videos and then finetuned on real videos using a small learning rate. Domain adversarial training seeks to find a shared latent space where data from the two domains are indistinguishable, usually through a minimax optimization of a GAN loss~\cite{goodfellow2014generative}.

\begin{figure}[t]
\begin{center}
   \includegraphics[width=0.9\linewidth]{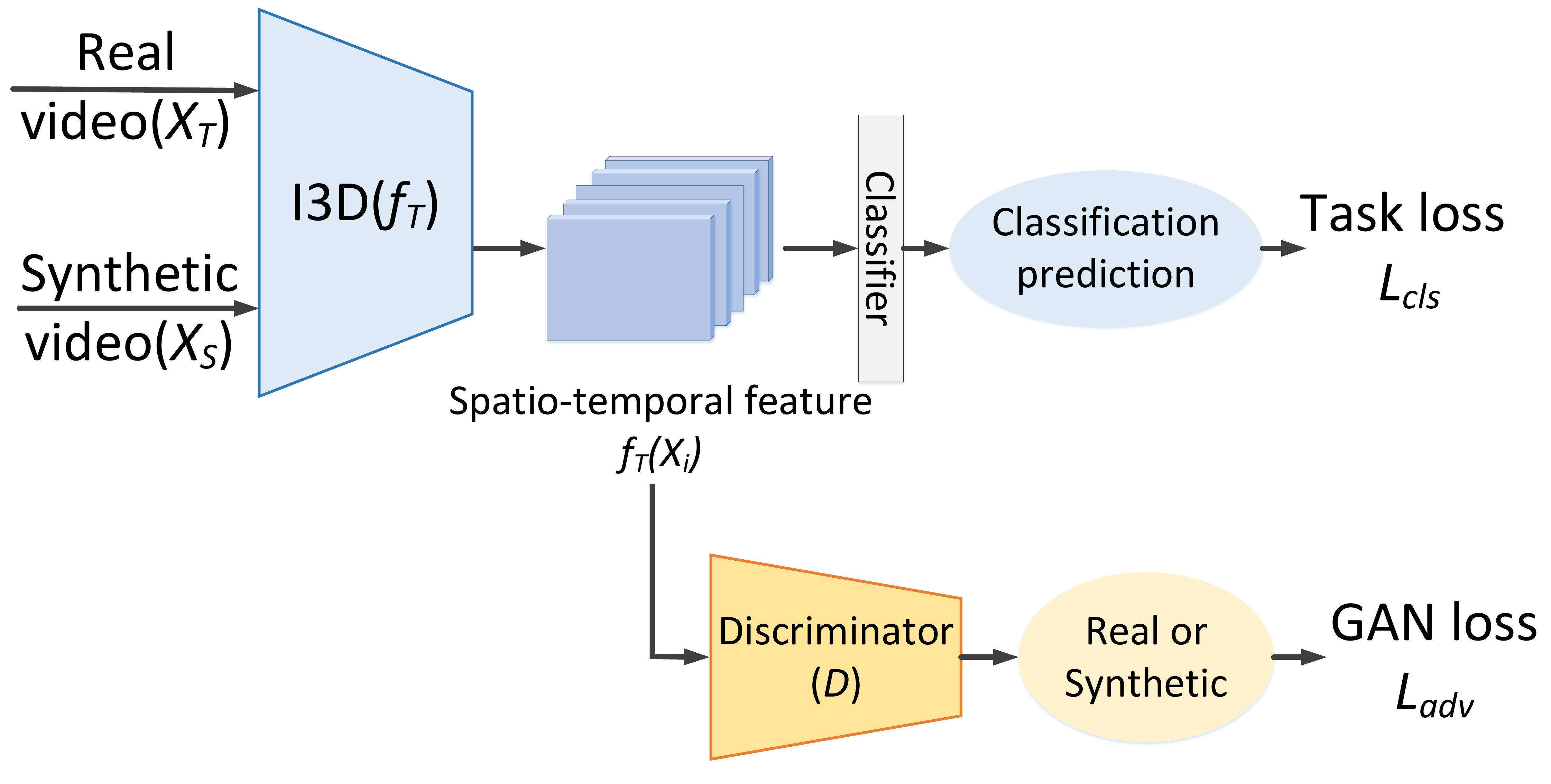}
\end{center}
   \caption{Domain adversarial training for the I3D network. A GAN loss is added to the training for alignment of synthetic and real domain in feature space. }
\label{fig:network_DA}
\end{figure}

Our domain adversarial training process is illustrated in Fig.~\ref{fig:network_DA}. Let $X_S$ and $X_T$ to be the distributions of source and target domains. $f_T$ denotes the module for extracting spatial temporal feature in an action recognition network $f$, e.g. the layers before the final $1\times1$ convolutional layer in a I3D network~\cite{carreira2017quo}. A discriminator $D$ is added to predict whether a feature is from source or target domain, which means maximizing the adversarial loss $\mathcal{L}_\text{adv}$, 
\begin{align}
\begin{split}
    \mathcal{L}_\text{adv}(f_T, D, X_s, X_t) = \mathbb{E}_{x_t\sim X_T}[\log D(f_T(x_t))] + \\
    \mathbb{E}_{x_s\sim X_S}[\log(1-D(f_T(x_s)))]
    \end{split}
\end{align}

The objective for the action recognition model is to bring the feature representation from the two domains closer as well as performing correct classification. Therefore, the final optimization is as follows,
\begin{equation}
    \min_{f} \max_{D} \mathcal{L}_{\text{cls}} + \lambda_{\text{adv}} \mathcal{L}_{\text{adv}} 
\end{equation}
where $\mathcal{L}_{\text{cls}}$ denotes the cross entropy loss for action classification on the two domains and $\lambda_{\text{adv}}$ is a weight for balancing the classification task and adversarial training. We perform training in an alternating manner, i.e. freezing the weight of $f$ when optimizing $D$ and vice versa. 

\subsection{Synthetic Data Generation Pipeline}

We developed two pipelines for data generation in our RSA framework to maximize the utilization of available 3D human resources. They are based on Unreal Engine and Blender respectively, each of which has unique properties. The first pipeline enables utilizing more high-quality commercial resources for building complex environments and human-object interaction. The second one is more suitable for controlling low-level rendering processing, such as controlling human meshes and rendering parameters.

For the Unreal Engine pipeline, we developed code for randomizing object textures, changing appearance of any 3D human model, and controlling the rotation of each bone on a human skeleton. A Python interface is provided which enables connecting the virtual environment with other learning algorithms. The whole pipeline can be released as a binary executable for reproducing and sharing our work. Our code for obtaining images and groundtruths is based on UnrealCV~\cite{qiu2017unrealcv}.  

For the Blender pipeline, human meshes from MakeHuman is utilized. The animation data captured from Microsoft Kinect~\cite{zhang2012microsoft} can be used by our pipeline to drive the human mesh in Blender. We also implemented a full pipeline to randomize other rendering factors.



\section{Experiments}

In this section we present the datasets used for evaluation, implementation details and experiment results showing the effectiveness of our proposed RSA framework. We also provide detailed analysis illustrating why our method works.

\subsection{Datasets for Evaluation}

\textbf{NTU RGB+D dataset~\cite{Shahroudy_2016_NTURGBD, Liu_2019_NTURGBD120}.} This is a large controlled human action dataset with variety of camera views, capture environments and human subjects. This dataset is collected from 106 distinct subjects and contains 114,480 sequences in total. There are 120 classes of human actions including daily one-person actions, two-human interactions and health-related activities. There are 32 different setups each of which has a different location and background. The actions are captured by three cameras at the same time covering the front, side and 45 degrees views. Depth and 3D skeletal data are also captured by Microsoft Kinetic. 

\textbf{Highly Disjoint Train/test split for NTU RGB+D.} The controlled real-world dataset enables us to evaluate the robustness of action recognition models to nuisance factors. We split this dataset so that train and test splits have no shared environments, human subjects and camera viewpoints. For environments, we use setup 7, 10, 11, 13, 21, 22, 23, 24, 25 for testing and the rest for training. We do not use the split for the cross-setup evaluation because the same environments appear in both train and test set in that split. For human subjects, we select human subject IDs for training following the cross-subject evaluation in \cite{Liu_2019_NTURGBD120}. For camera viewpoints, we only choose the camera view 1 (C001) for training and test on others. In this work, we focus on a subset of 31 classes which do not require reasoning about human-object interactions. This results in a training set of 3950 videos and a test set of 2670 videos. 

\begin{figure}[t]
\begin{center}
   \includegraphics[width=\linewidth]{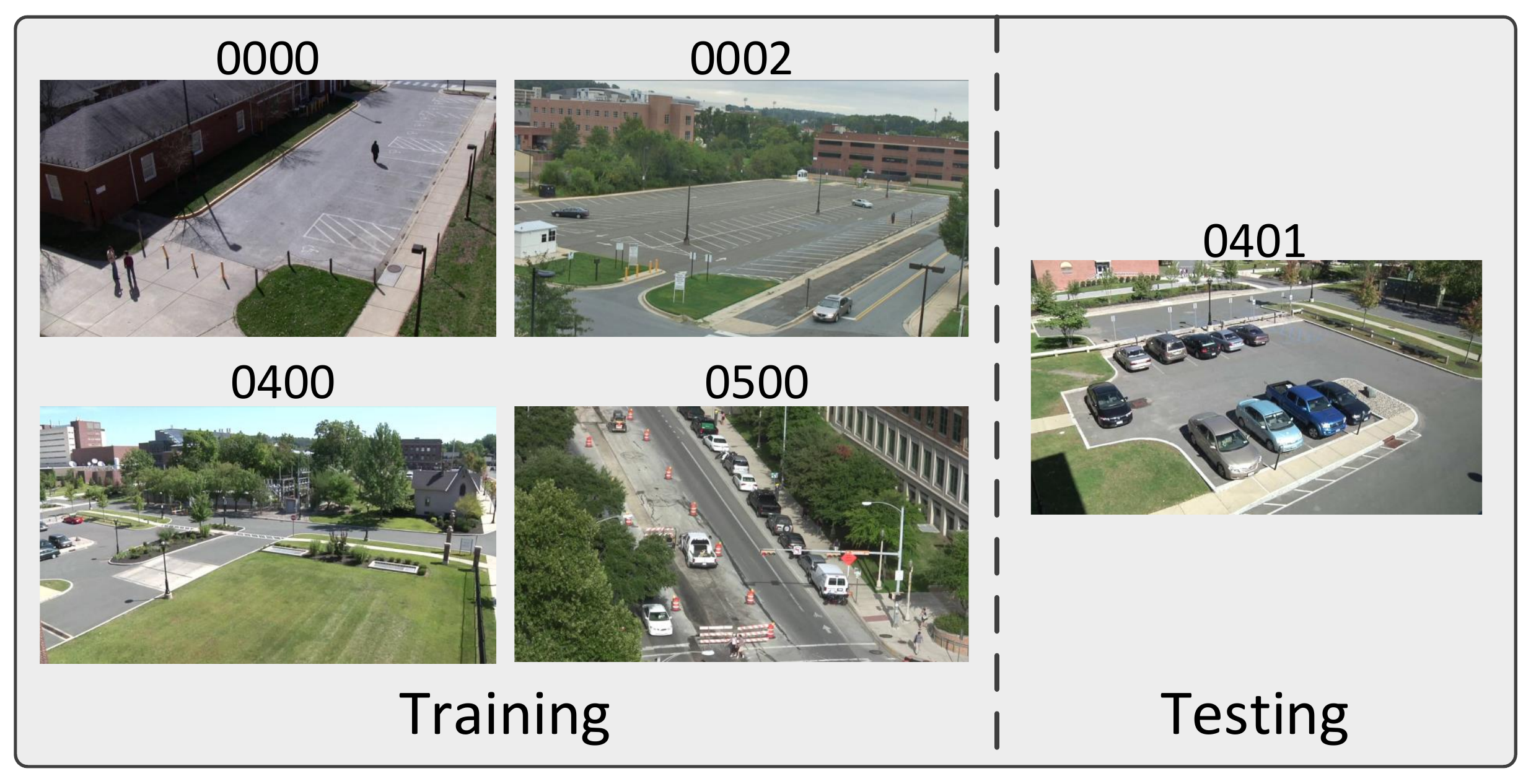}
\end{center}
   \caption{The Leave-One-Scene-Out (LOSO) setting for the five scenes in VIRAT dataset; the left four scenes are used for training and the right one is left out for testing.}
\label{fig:loso}
\end{figure}

\textbf{VIRAT dataset~\cite{oh2011large}.} This is a dataset for video surveillance containing videos with cluttered backgrounds and various camera viewpoints, including 19 classes of human actions and human-vehicle interactions. The videos are collected at multiple sites distributed throughout the USA, including five scenes named 0000, 0002, 0400, 0401 and 0500. The scenes are shown in Fig.~\ref{fig:loso}, the videos in these scenes share the same actions but have different backgrounds and viewpoints. VIRAT dataset is challenging due to 1) low resolution actions, i.e. each action only occupies a small region in the whole frame; 2) only limited data available for training. We focus on the task of classifying video clips that has already been cropped both spatially and temporally using ground truth annotation.

\textbf{Highly Disjoint Train/test split for VIRAT.} The differences between five scenes enable us to create new splits suitable for evaluating the robustness of action recognition model to nuisance factors. We split the dataset based on Leave-One-Scene-Out (LOSO) criterion, in which we use videos from one scene for testing and the rest of training. In our experiment, we leave $0401$ out in order to keep the balance between training and testing set. In the new split, we focus on six typical yet challenging classes of activities in VIRAT: Opening, Open Trunk, Entering, Exiting, Close Trunk and Closing. The training set contains 534 video clips and testing set contains 363 videos clips.

\subsection{Implementation Details}

\noindent\textbf{Network architecture.} For all experiments, we use the I3D network~\cite{carreira2017quo} as our base action recognition network which is initialized by inflating weights from 2D Inception network pretrained on ImageNet~\cite{deng2009imagenet}. We rescale the shorter side of all input videos to 256 pixels. For training, we randomly sample 32 consecutive frames temporally and crop a $224\times224$ patch from each frame spatially. Random left-right flip is applied as 2D data augmentation. At test time, we spatially sample three patches evenly along the longer side of the video and temporally sample three clips evenly from the full video. For implementation of domain adversarial training, we use a simple discriminator with three 3D convolutional layers each followed by a 3D BatchNorm Layer~\cite{ioffe2015batch} and a Leaky ReLU activation. 

\noindent\textbf{Human Motion Model Acquisition.} For NTU RGB+D dataset, the 3D skeleton data is already provided which is calculated by Kinect. We find the predefined skeleton of Kinect, calculate the local rotations of all the joints in each frame and write the animations in BVH file format. We then import the generated BVH files into Blender and use these animations to drive 3D Human models. For VIRAT dataset, we use a low-cost motion capture device to collect a set of human motions for each action class.

\begin{figure}[t]
\begin{center}
   \includegraphics[width=\linewidth]{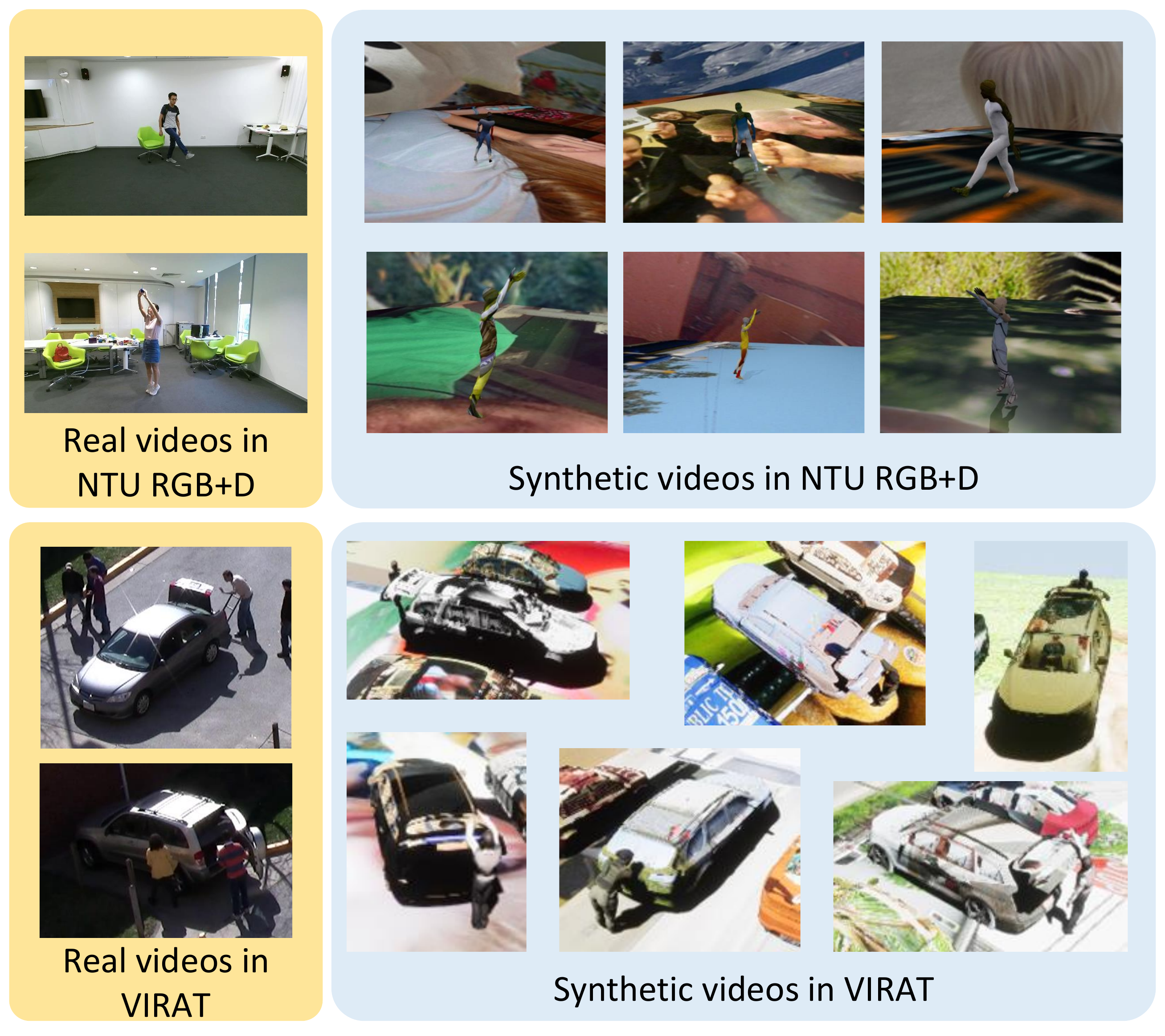}
\end{center}
   \caption{Comparison of real video frames and our synthetic video frames. The top two rows are for NTU RGB+D dataset and the bottom two rows are for VIRAT dataset. The examples in NTU RGB+D are from \textit{kicking something} and \textit{shoot at basket}, in VIRAT are from \textit{Opening Trunk}.}
\label{fig:com_diva}
\end{figure}

\subsection{Experimental Results on NTU RGB+D}

In this section we evaluate RSA on the NTU RGB+D dataset. We generate a total of 14631 synthetic videos, which is about 4 times larger than the real training set, to augment the real training set. 
We compare three methods for learning from our generated synthetic data: 1) direct joint training (\textbf{Joint training}) as our baseline; 2) pretraining on synthetic videos and finetuning on real videos (\textbf{RSA-F}); and 3) domain adversarial training (\textbf{RSA-DA}). 

As shown in Table~\ref{tab:ntu_res}, RSA improves the action recognition accuracy compared to training purely on real videos. The classification accuracy improves from $83.93\%$ to $88.69\%$ ($+4.76\%$) with finetuning, and to $85.02\%$ ($+1.09\%$) with adversarial training. This suggests that with RSA the state-of-the-art I3D model achieves increased robustness to nuisance factors. Due to the reality gap, model trained on synthetic data only achieve an accuracy of $47.57\%$. Direct joint training performs worse than real-only, showing that handling the reality gap carefully is important. 

In the rest of this section, we provide more experiments for better understanding of our framework, including: effects with reduced real videos, ablation studies on the effect of individual nuisance factor and visualization of learned features from different ways of handling the reality gap.

\begin{table}
\begin{center}
\begin{tabular}{l|c}
\hline
Model & Accuracy \\
\hline
Real-only & 83.93  \\
\hline
Synthetic-only &  47.57 \\
\hline
Joint training &  82.92 \\
\hline
RSA-F &  \textbf{88.69} \\
RSA-DA & 85.02  \\
\hline
\end{tabular}
\end{center}
\caption{\label{tab:ntu_res} Results on NTU RGB+D dataset in accuracy (\%). }
\end{table}

\begin{figure}
  \includegraphics[width=\linewidth]{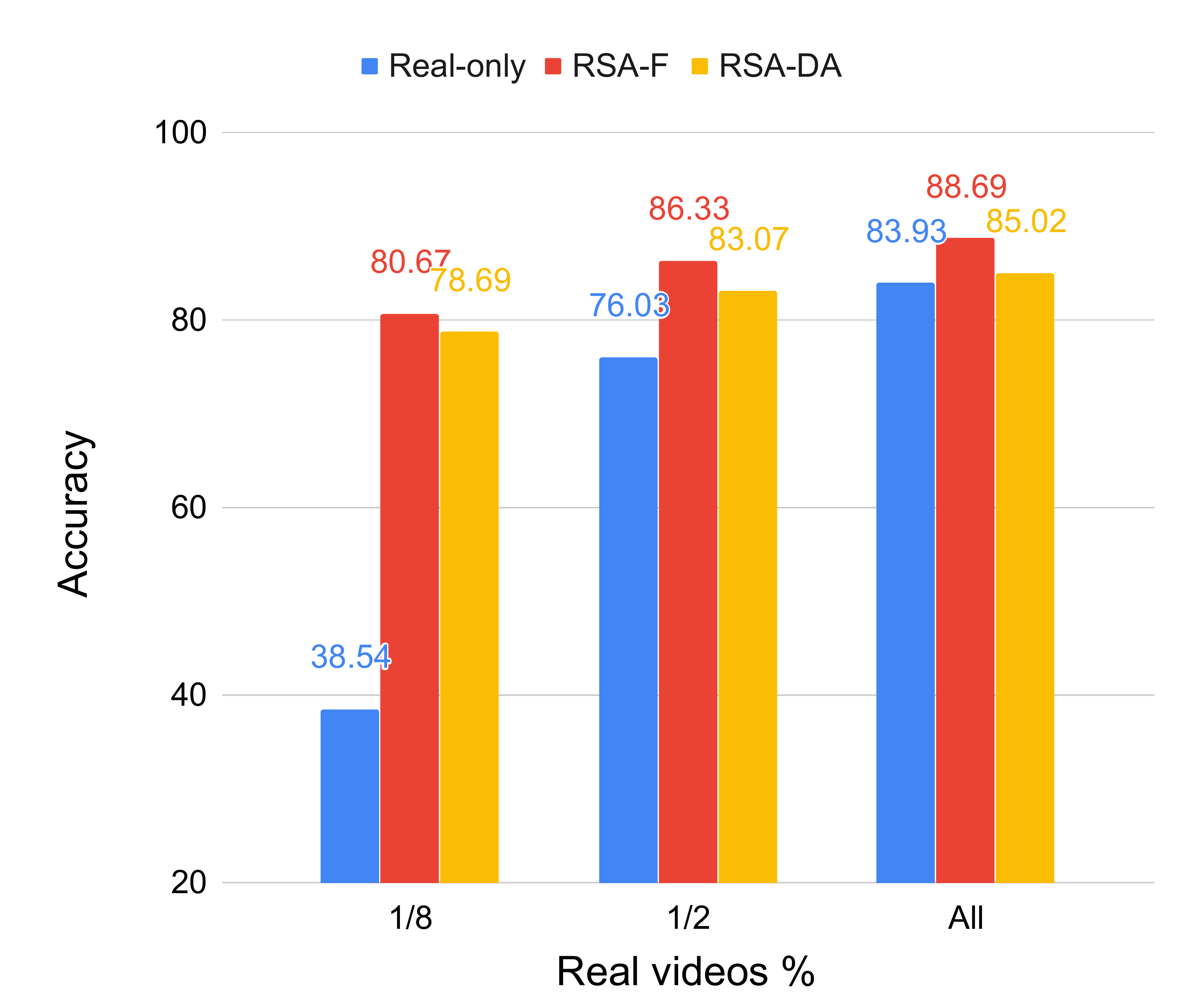}
  \caption{Performance boost with RSA on NTU RGB+D dataset using less real data. }
  \label{fig:data_ratio}
\end{figure}

\begin{table*}[]
\begin{tabular}{|l||c|c|c||c|c|c||c|c|c|}
\hline
\multirow{2}{*}{Model} & \multicolumn{3}{c||}{All Real Data} & \multicolumn{3}{c||}{1/2 Real Data} & \multicolumn{3}{c|}{1/8 Real Data} \\ \cline{2-10} 
 & Subject & Background & Camera & Subject & Background & Camera & Subject & Background & Camera \\ \hline
Real-only & 94.50 & 93.11 & 87.32 & 92.00 & 86.48 & 82.28 & 64.04 & 51.41 & 54.14 \\ \hline
RSA-F & \textbf{95.48} & \textbf{94.08} & \textbf{90.27} & \textbf{93.17} & \textbf{91.86} & \textbf{87.77} & \textbf{87.92} & \textbf{83.33} & \textbf{81.06} \\
RSA-DA & 93.74 & 90.66 & 88.18 & 92.06 & 89.52 & 85.47 & 87.48 & 82.08 & 79.13 \\ \hline
\end{tabular}
\caption{\label{tab:nf_ablation} Results for individual nuisance factors on NTU RGB+D dataset. We evaluate model robustness to human subject changes (Subject), different background (Background) and different camera viewpoints (Camera). Experiments are performed with different amount of real videos for training. }
\end{table*}

\textbf{Effectiveness with fewer training videos.} To evaluate how much improvement RSA can achieve with less real training data, we perform experiments on NTU RGB+D with only a subset of the real videos available for training. As shown in Fig.~\ref{fig:data_ratio}, consistent performance boosts are achieved by adding RSA. As the number of labeled real data decreases, I3D performance drops as expected due to an increased possibility of overfitting to training data. With the help of our proposed method, the same level performance (accuracy down $8\%$) is achieved with only 1/8 of all real training data, whereas real-only performance drops significantly (accuracy down $45\%$). This shows that our framework can significantly reduce the number of labeled real videos need to achieve good performance.

\textbf{Improvement for individual nuisance factor.} To study how well our model works for each nuisance factor, we create new test sets varying one nuisance factor at a time. For example, to evaluate the robustness to different human subjects, we keep the camera view and setup the same as training and choose videos performed by human subjects in the test set. Results are shown in Table~\ref{tab:nf_ablation}. Our method improves the robustness of all three nuisance factors especially when training data is insufficient. Among the three nuisance factors evaluated, the camera viewpoint results to be the least robust one for I3D, which can be greatly improved by our method.  

\begin{figure*}[t]

\centering
    \begin{subfigure}[b]{0.3\textwidth}
         \centering
         \includegraphics[width=\textwidth]{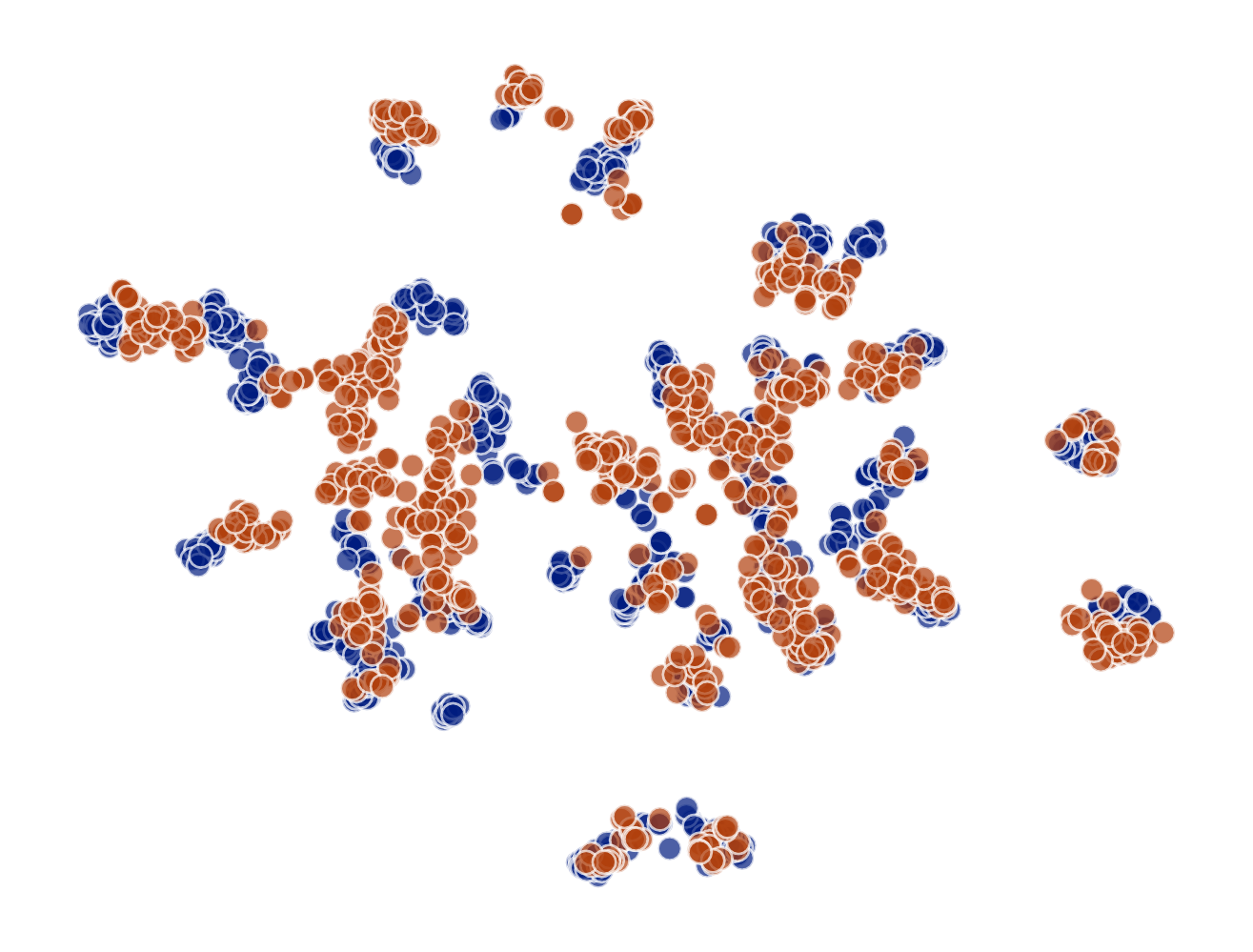}
         \caption{Joint Training\label{fig:feature_joint_training}}
     \end{subfigure}
     \begin{subfigure}[b]{0.3\textwidth}
         \centering
         \includegraphics[width=\textwidth]{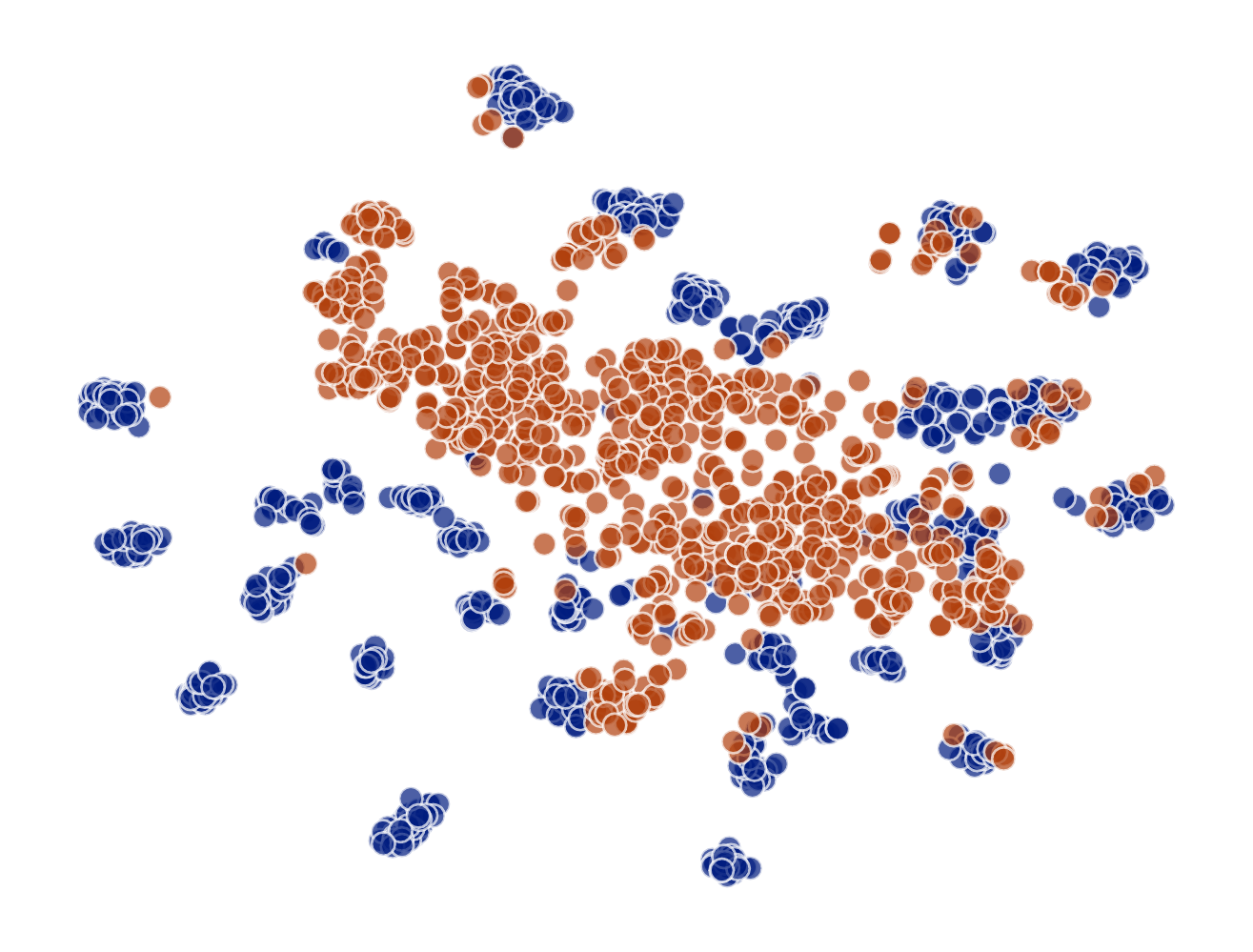}
         \caption{RSA-F\label{fig:feature_finetune}}
     \end{subfigure}
     \begin{subfigure}[b]{0.3\textwidth}
         \centering
         \includegraphics[width=\textwidth]{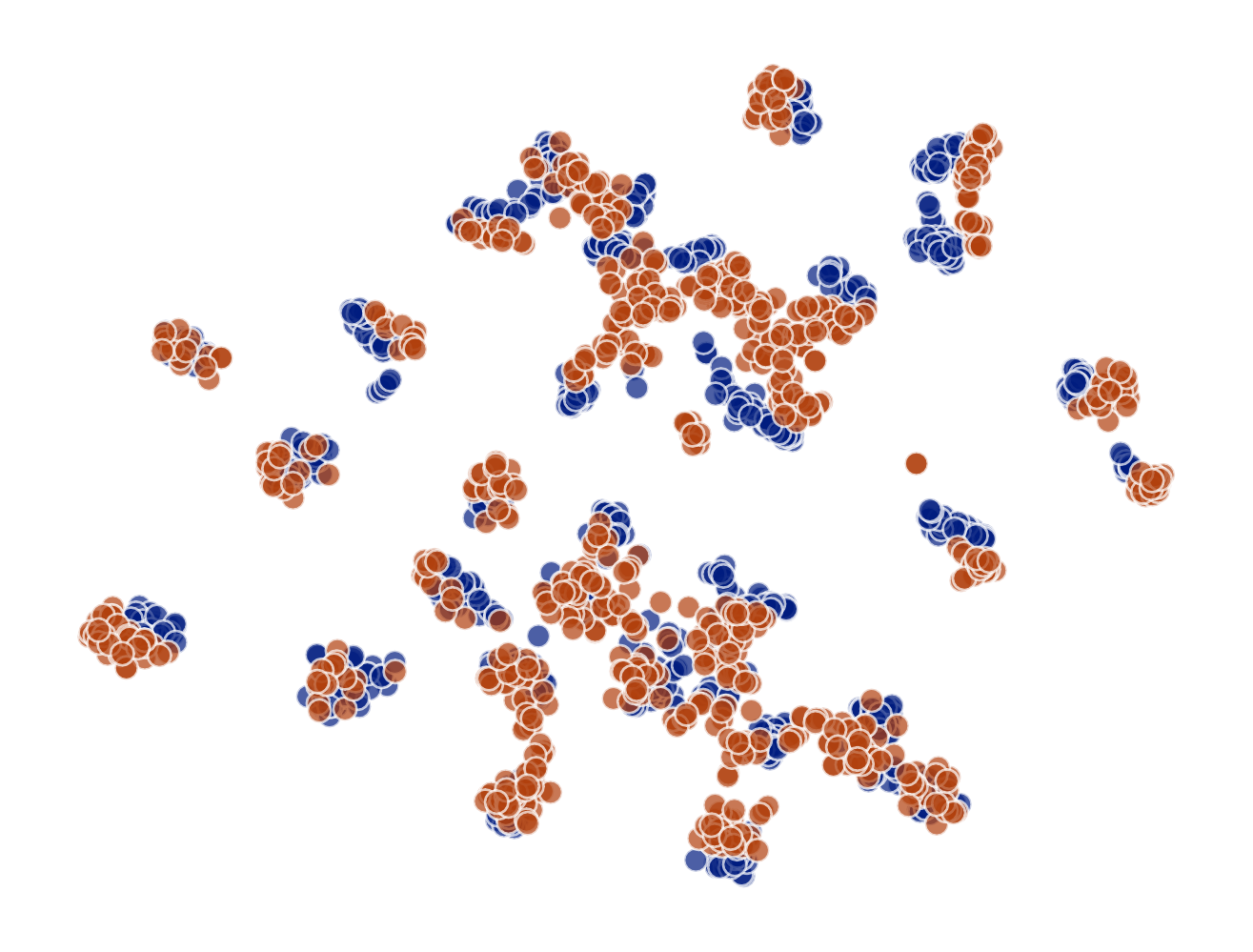}
         \caption{RSA-DA\label{fig:feature_domain_adaptation}}
     \end{subfigure}
\caption{Feature embedding in 2D space using t-SNE for models trained with different strategy for handling the reality gap. Blue: Real domain; Red: Synthetic domain. (Better viewed in color.) (a) Joint training with synthetic and real data achieves moderate alignment of the two domains but the learned features are not very discriminative. (b) Finetuning on real data results in high discriminative ability on real domain. (c) Domain adversarial training aligns the two domains better than using joint training while keeping the features in separate clusters. }
\label{fig:feature_embed}
\end{figure*}

\textbf{Visualization for Handling the Reality Gap.} We visualize the feature distributions of data from both domains learned by different models in 2D space using t-SNE~\cite{maaten2008visualizing}. Fig.~\ref{fig:feature_embed} shows feature embedding for models trained with direct joint training, finetuning and domain adversarial training. Compared with joint training, domain adversarial training achieves better alignment for synthetic and real domain while learning more discriminative features, i.e. features are in separate clusters. This can explain why domain adversarial training achieves better results than joint training. As shown in Fig.~\ref{fig:feature_finetune}, after finetuning on real videos, I3D focuses on the real domain and is no longer able to handle synthetic data well. 

\subsection{Experimental Results on VIRAT}
In this section, we evaluate RSA on VIRAT dataset using the same three methods for learning from synthetic data as on NTU RGB+D dataset. VIRAT is quite different with NTU RGB+D; it contains less data and is more difficult, the human and vehicle in the video are far from the camera and the surveillance video itself is blurred thus hard to recognize the activity; the latter is collected indoor and the action is explicit thus easy to recognize. Therefore, the accuracy of model on VIRAT is generally lower than that on NTU RGB+D. We experiment on VIRAT to show our proposed method's ability of handling difficult dataset, for which we generated 7582 synthetic videos (14 times the size of real training data). 

As is shown in Table~\ref{tab:virat_res}, RSA greatly improves the accuracy compared with training only on real. While the joint training baseline hardly improves the accuracy, finetuning lifts the accuracy from $46.01\%$ to $54.82\%$ (+$8.81\%$) and domain adversarial training lifts the accuracy from $46.01\%$ to $55.10\%$ (+$9.09\%$). We can see domain adversarial training works better than finetuning, unlike the results on NTU RGB+D, which is probably because the number of real data on VIRAT is insufficient for finetuning. 

In the following parts, we provide experiments to show the influence of nuisance factors and detailed analysis of the improvement brought by our method.

\begin{table}
\begin{center}
\begin{tabular}{l|c}
\hline
Model & Accuracy \\
\hline
Real-only & 46.01  \\
\hline
Synthetic-only & 33.61  \\
\hline
Joint training &  46.28 \\
\hline
RSA-F &  54.82 \\
RSA-DA & \textbf{55.10}  \\
\hline
\end{tabular}
\end{center}
\caption{\label{tab:virat_res} Results on VIRAT dataset in accuracy (\%). }
\end{table}

\textbf{Influence of nuisance factors}
We did experiments on original train/test splits of VIRAT and compare with the LOSO splits to quantitatively show the effect of nuisance factors. The training and testing sets in the original splits share similar background, viewpoints and vehicle appearance, while in LOSO splits the two sets have little overlap on these factors. Results are in Table~\ref{tab:virat_nui}. The result shows that dramatic variation of the nuisance factors poses a challenge to state-of-the-art action recognition network as the I3D performance drops from $52.97\%$ to $46.01\%$ ($-6.96\%$).

\textbf{Improvement on each class}
We evaluate the performance for each action class using F1 score metric, which is defined as the harmonic mean of precision and recall. The results are shown in Table~\ref{tab:virat_each_real}. Domain adversarial training brings improvement to all the classes while finetuning brings improvement to most of the classes except \textit{Open Trunk}. Great improvement is achieved on the \textit{Entering} class which is difficult for model trained on real only. Model trained using joint training achieves improvement on some classes while performing terribly on \textit{Exiting}. 

We also find an interesting fact that the model trained with synthetic data only performs even better than real data trained model on some classes, such as $Entering$ and $Opening$. This suggests that the model can learn useful features from synthetic data with randomized nuisance factors despite the reality gap.

\begin{table}
\begin{center}
\setlength{\tabcolsep}{1.3mm}{
\begin{tabular}{l|c|cccccc}
\hline
Model & \rotatebox{90}{\textit{Average}} & \rotatebox{90}{\textit{Closing}} & \rotatebox{90}{\textit{Close Trunk}} & \rotatebox{90}{\textit{Entering}} & \rotatebox{90}{\textit{Exiting}} & \rotatebox{90}{\textit{Open Trunk}} & \rotatebox{90}{\textit{Opening}} \\
\hline
Real-only & 44.4 & 55.7 & 38.5 & 23.2 & 52.1 & 40.8 & 39.1 \\
\hline
Synthetic-only & 32.3 & 33.7 & 27.8 & 38.0 & 24.0 & 18.9 & 41.6 \\
\hline
Joint training & 44.8 & 53.0 & 41.9 & 27.8 & 30.4 & \textbf{47.8} & 51.9 \\
\hline
RSA-F & 51.8 & 61.2 & 39.1 & 44.4 & \textbf{60.6} & 40.0 & \textbf{58.1} \\
RSA-DA & \textbf{52.0} & \textbf{65.8} & \textbf{43.6} & \textbf{46.0} & 56.7 & 42.4 & 54.0 \\
\hline
\end{tabular}}
\end{center}
\caption{\label{tab:virat_each_real} F1 score of each class of activity on VIRAT real test set in LOSO setup. }
\end{table}

\begin{table}
\begin{center}
\begin{tabular}{l|c}
\hline
Split & Accuracy \\
\hline
Original &  52.97 \\
LOSO & 46.01  \\
\hline
\end{tabular}
\end{center}
\caption{\label{tab:virat_nui} Comparison of classification accuracy for original and LOSO split of VIRAT dataset. I3D network used. }
\end{table}

\section{Conclusion}

In this work, we propose the RSA framework which augments real-world videos with randomized synthetic data for training more robust human action recognition networks. Domain gap between synthetic and real data is carefully handled so that model can learn transferable features. We demonstrate on two real-world datasets that our method boosts the performance of the state-of-the-art I3D network for human action recognition and improves model robustness to nuisance factors. Furthermore, we show that our method can reduce real-world labeled data needed for achieving good performance. This can potentially save a huge amount of human labors for collecting and annotating large-scale video datasets. In our future work, we plan to extend our model to simulate more complex human-object interactions and explore the potential of using the free extra groundtruths from simulation, e.g. depth and segmentation masks, for better action recognition models.

\noindent\footnotesize\textbf{Acknowledgement:} This work is supported by IARPA via DOI/IBC contract No. D17PC00342. We thank Tae Soo Kim, Mike Peven, Jiteng Mu and Jialing Lyu for discussions.

{\small
\bibliographystyle{ieee_fullname}
\bibliography{egbib}
}

\end{document}